\documentclass{article}

\PassOptionsToPackage{numbers, compress}{natbib}

\usepackage[preprint]{neurips_2026}

\usepackage[utf8]{inputenc} 
\usepackage[T1]{fontenc}    
\usepackage{hyperref}       
\usepackage{url}            
\usepackage{booktabs}       
\usepackage{amsfonts}       
\usepackage{nicefrac}       
\usepackage{microtype}      
\usepackage{xcolor}         
\usepackage{graphicx}

\usepackage{amsmath}
\usepackage{amssymb}
\usepackage{mathtools}
\usepackage{amsthm}

\usepackage{xcolor}
\usepackage{mdframed}

\definecolor{purplefill}{RGB}{238,237,254}
\definecolor{purplerule}{RGB}{83,74,183}
\definecolor{purpletext}{RGB}{60,52,137}
\definecolor{burgundyfill}{RGB}{250,238,238}
\definecolor{burgundyrule}{RGB}{155,40,55}
\definecolor{burgundytext}{RGB}{130,25,40}
\usepackage[capitalize,noabbrev]{cleveref}
\theoremstyle{plain}

\theoremstyle{definition}

\theoremstyle{remark}

\usepackage{nicefrac}       
\usepackage{physics}        

\usepackage{xcolor}%
\usepackage{pdflscape}
\usepackage{multirow}
\usepackage{subcaption}
\usepackage{tabularx}
\usepackage{afterpage}

\usepackage[normalem]{ulem} 
\usepackage{blindtext}%
\usepackage[textsize=scriptsize]{todonotes}%
\presetkeys{todonotes}{fancyline}{}

\usepackage{url}%

\usepackage{enumitem} 

\usepackage{xparse} 
\usepackage{float}
\usepackage{placeins}

\newcommand{\Exp}{\mathrm{Exp}}
\newcommand{\Log}{\mathrm{Log}}

\usepackage{algorithm}
\usepackage[noend]{algpseudocode}

\usepackage{amsmath}
\usepackage{amssymb}
\usepackage{bm}
\usepackage{mathtools}

\def\veta{{\bm{\eta}}}
\def\vzeta{{\bm{\zeta}}}
\def\vOmega{{\bm{\Omega}}}

\def\mA{{\bm{A}}}
\def\mB{{\bm{B}}}

\def\mG{{\bm{G}}}

\def\mH{{\bm{H}}}
\def\mI{{\bm{I}}}

\def\mO{{\bm{O}}}

\def\mQ{{\bm{Q}}}
\def\mR{{\bm{R}}}

\def\mU{{\bm{U}}}
\def\mV{{\bm{V}}}
\def\mW{{\bm{W}}}

\DeclareMathAlphabet{\mathsfit}{\encodingdefault}{\sfdefault}{m}{sl}
\SetMathAlphabet{\mathsfit}{bold}{\encodingdefault}{\sfdefault}{bx}{n}

\def\gG{{\mathcal{G}}}

\def\gM{{\mathcal{M}}}

\def\sR{{\mathbb{R}}}

\DeclareSymbolFont{bbold}{U}{bbold}{m}{n}
\DeclareSymbolFontAlphabet{\mathbbold}{bbold}

\DeclareMathOperator{\GL}{GL}

\begin{document}

\newcommand{\papertitle}{%
Generalizing the Geometry of Model Merging Through Fr\'echet Averages
}%
\newcommand{\shortpapertitle}{%
Generalizing the Geometry of Model Merging
}%

\title{\papertitle}

\author{Marvin F.~da Silva$^{1,2}$\thanks{Corresponding Author. marvinf.silva@dal.ca}
\quad
Mohammed Adnan$^{2,3}$
\quad
Felix Dangel$^{4,5}$
\quad
Sageev Oore$^{1,2}$ \\
  $^{1}$Faculty of Computer Science, Dalhousie University \\
$^{2}$Vector Insitute for Artificial Intelligence \\
$^{3}$ Schulich School of Engineering, University of Calgary \\
$^{4}$ Department of Computer Science and Software Engineering, Concordia University \\
$^{5}$ Mila - Quebec AI Institute \\
}
\maketitle 
\vskip -0.3in



\begin{abstract}
Model merging aims to combine multiple models into one without additional training. Naïve parameter-space averaging can be fragile under architectural symmetries, as their geometry does not take them into account. In this work, we argue that not only the geometry but also the averaging procedure itself must be symmetry-invariant to achieve symmetry-aware merges. Consequently, we propose a general solution: merging as Fr\'echet averaging, i.e.\,selecting parameters that minimize a sum of geodesic distances on an appropriate manifold. In this view, the key design choice is the overall geometry, i.e.,\,the choice of metric, manifold, and distance approximation, that determines what it means for two models to be “close.”
We show that Fr\'echet averaging, combined with simplifying assumptions, contains Fisher merging. Building on this, we examine the particular case of low-rank adapters (LoRA), whose symmetries induce a distinct geometry: that of a quotient manifold. We outline limitations of current LoRA merging methods, propose a practical algorithm for this setting, and support the effectiveness of our method with empirical results.
\end{abstract}

\section{Introduction}
Model merging aims to combine multiple trained networks into a single model that preserves the union of their capabilities while avoiding additional end-to-end training.
This objective is increasingly salient because modern workflows routinely produce \emph{families} of related models: different random seeds and hyperparameters, domain- or task-specialists, safety patches, preference updates, and personalized variants.
Existing approaches broadly fall into two classes:
\emph{(1) Data-dependent} methods require data and/or gradients at merge time; examples include Fisher-weighted merging~\citep{matena2022fisher}, RegMean~\citep{jin2023regmean}, and MaTS~\citep{tam2024mats}, which are strong when data is available but less applicable when it is proprietary or costly to process.
\emph{(2) Data-free} (weight-only) methods operate purely in parameter space and remain deployable with only model checkpoints; this includes model soups~\citep{wortsman2022soups}, Task Arithmetic~\citep{ilharco2023taskarith}, and TIES~\citep{yadav2023ties}. We focus on this class throughout this work.

\begin{figure*}[!t]
  \centering
  \begin{subfigure}[t]{0.495\textwidth}
    \centering
    \includegraphics[width=\textwidth]{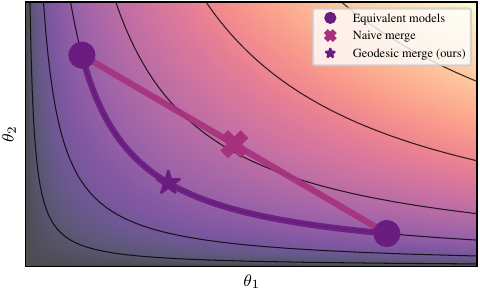}
  \end{subfigure}
  \hfill
  \begin{subfigure}[t]{0.495\textwidth}
    \centering
    \includegraphics[width=\textwidth]{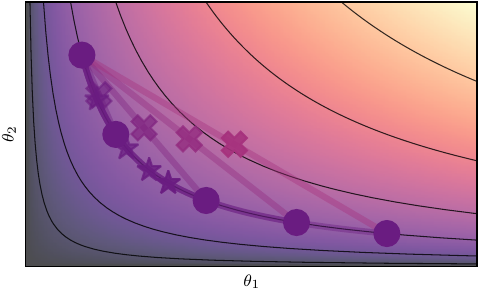}
  \end{subfigure}
  \caption{Visual comparison of different model merging approaches, highlighting failure scenarios due to symmetry unawareness.
    \emph{Left:} Naive averaging steps off the orbit because it uses the wrong geometry.
    \emph{Right:} Naive merging is ambiguous and lands on different orbits depending on parameterization.
    Geodesic merging always stays on the same orbit as it uses symmetry-invariant averaging.}
  \label{fig:visual_abstract}
  \vspace{-2ex}
\end{figure*}

\textbf{Misalignment complicates adapter merging.}
When models are full-rank fine-tuned from the same pretrained initialization, these operations can yield strong merged performance without further training~\citep{wortsman2022soups,ilharco2023taskarith,matena2022fisher,yadav2023ties}.
\textit{However, this success does not reliably transfer to models produced via parameter-efficient fine-tuning (PEFT).}
Low-Rank Adaptation (LoRA)~\citep{hu2022lora} constructs specialists by learning low-rank updates to selected matrices, yet weight-only merge rules that behave benignly for full-rank fine-tunes can degrade sharply when applied to LoRA specialists, even when 
specialists share the same base model \citep{stoica2025knots}.
\citet{stoica2025knots} propose that this gap is fundamentally about \emph{alignment of task updates}.
Comparing pairwise centered kernel alignment (CKA) of representations attributable to fine-tuning updates alone, they find that full-rank fine-tuned models exhibit 
high CKA, while LoRA counterparts show substantially lower CKA, suggesting 
different LoRA task-updates process inputs through \emph{misaligned} subspaces.
This misalignment correlates with destructive interference under naive addition or averaging and motivates an explicit alignment stage: express task-updates in a shared basis (e.g., via an SVD construction), then apply merge operators (e.g., Task Arithmetic or TIES) in the aligned coordinate system. Alignment, however, raises a deeper question:

\textbf{What exactly are we trying to align?}
Neural network parameters are not canonical coordinates for functions, and modern architectures have large symmetry groups under which distinct parameter settings implement exactly the same input-output map.
Classic examples include neuron permutations \citep{nielsen1990} and rescaling symmetries in positively homogeneous networks \citep{dinh2017sharp}.
Transformers exhibit richer continuous symmetries: attention blocks admit high-dimensional $\GL(h)$ symmetries that generate directions along which the network function (and loss) is unchanged \citep{dasilva2025hide,zhang2025beyond}.
These symmetries can obscure both alignment analyses and procedures.
For low-rank updates, the usual factorization introduces a $\GL(r)$ symmetry $(G,H)\sim \smash{(GA^{-1},HA^\top)}$ that preserves the product $\smash{GH^\top}$, so apparent misalignment can be partly a coordinate artifact. \citet{wang2025modelunmergingmakingmodels} found that applying one of these $\GL(r)$ transformations, the performance of merged models collapses.
This suggests that 
robust alignment requires distances and averages intrinsic to the space of functions (or updates), not to an arbitrary parameterization.

\textbf{In summary.} ~\cref{fig:visual_abstract} illustrates the core geometric failure mode behind many weight-only merges under symmetry: when multiple parameter settings represent the same underlying model, Euclidean averaging can (i) step off the orbit because it measures “closeness” in the wrong geometry, and (ii) become ambiguous because different but equivalent parameterizations yield different merges.

\textbf{Our solution: averaging on Riemannian manifolds.} A principled way to do this averaging is to treat the parameter space as a \emph{quotient} by the relevant symmetry group and define geometric notions intrinsically on that quotient.
\citet{dasilva2025hide} develop this viewpoint for transformers, arguing that symmetry can invalidate Euclidean geometric measures and proposing symmetry-corrected objects on the quotient manifold, mapped back to parameter space via horizontal lifts.

We ask whether the geometric methodology can be elevated to a \emph{constructive} principle for merging:
\vspace{-0.05in}
\begin{mdframed}[
  leftline=true, rightline=false,
  topline=false, bottomline=false,
  linewidth=3pt, linecolor=burgundyrule,
  backgroundcolor=burgundyfill,
  innerleftmargin=10pt, innerrightmargin=10pt,
  innertopmargin=8pt, innerbottommargin=8pt,
  skipabove=4pt, skipbelow=4pt
]
Can we formulate model merging as averaging
on a Riemannian manifold whose geometry respects parameter symmetries to
eliminate symmetry-related failures?
\end{mdframed}
\vspace{-0.05in}
We develop \textbf{GeoMerge}, a geometric framework that treats merging as computing a \emph{Fr\'echet mean} in a chosen Riemannian representation space.
In this view, a merge method amounts to geometric choices: (a) an appropriate parameter manifold; (b) an appropriate metric; and (c) any equivalence on the parameters we wish to choose.
Weight-only merging uses representations and metrics computable without data, while data-dependent merging uses metrics estimated from activations, gradients, or curvature, linking merging to information geometry.
Our contributions are:
\begin{itemize}[leftmargin=1em]
    \item \textbf{\Cref{sec:prelims,sec:geomerge}:} We cast model merging as Fr\'echet averaging on a Riemannian manifold and argue that, under architectural symmetries, the correct objects live on a quotient manifold, yielding invariance with respect to gauge choices \citep{pennec2006intrinsic,boumal2023introduction}.
    
    \item \textbf{\Cref{sec:relations}}: We connect importance-weighted merging, specifically Fisher merging \citep{matena2022fisher}, to Fr\'echet objectives under information-geometric metrics, highlighting their symmetry non-invariance despite employing a refined geometry, due to their symmetry-unaware average.

    \item \textbf{\Cref{sec:high_rank,sec:experiments}:} Motivated by $\GL$ symmetries in low-rank updates \citep[LoRA,][]{hu2022lora} and attention \citep{dasilva2025hide}, we develop quotient-compatible primitives that yield symmetry-corrected merging algorithms for LoRA factors. We provide a quotient-aware lifting scheme for embedding low rank adapters into a higher dimensional space to assist in merging.
    We illustrate geodesic merging analytically for a small toy model (\Cref{sec:toy-problem}), and as proof of concept, we scale the computational algorithm to LoRA adapters for larger models (ViT-B/32 in ~\cref{sec:experiments}, Llama 3 8B in~\cref{sec:nli}).
\end{itemize}

\subsection{Other Related Work}
\textbf{Weight averaging and basin structure.}
Model soups \citep{wortsman2022soups} show that averaging fine-tuned checkpoints can improve accuracy without inference-time overhead.
A common explanation invokes flatness and (approximate) linear mode connectivity \citep{garipov2018mode,draxler2018essentially}, which makes Euclidean averaging benign when solutions live in one convex-like region.
GeoMerge does not contradict this; rather, it asks what happens when Euclidean geometry is \emph{not} the right notion of closeness---a question that becomes especially pressing in the presence of symmetries.

\textbf{Symmetries, quotient geometry, and deep learning.}
Weight-space symmetries are known to invalidate naive Euclidean notions of sharpness and flatness \citep{dinh2017sharp}.
\citet{dasilva2025hide} study the high-dimensional \(\GL(h)\) symmetries in transformers’ attention heads, and develop symmetry-corrected sharpness measures on quotient manifolds \citep{boumal2023introduction}.
We extend their recipe by studying Riemannian averaging operators and target \emph{model merging} rather than sharpness.

\textbf{KnOTS and LoRA.} 
KnOTS \citep{stoica2025knots} improves LoRA merging via joint SVD-based transformations to better align low-rank updates before applying existing merge rules.
GeoMerge complements this; it lets us see SVD-based alignment as a particular gauge fixing for $\GL(r)$ symmetries of low-rank factorizations, selecting comparable representatives of the same underlying updates before averaging.

\section{Preliminary Definitions, Notation \& Math}
\label{sec:prelims}
We consider $T$ task fine-tuned adapters $\{\theta_t\}_{t=1}^T$ derived from a common base model. We will define a geometric merge on a (possibly quotient) parameter manifold $\mathcal{M}$.

\subsection{General Remarks}

\textbf{Riemannian manifolds, distance, and Exp/Log.}
A \emph{Riemannian manifold} is a pair $(\mathcal{M}, g)$, where $\mathcal{M}$ is a smooth manifold
and the metric $g$ assigns to each $p \in \mathcal{M}$ an inner product
$g_p : T_p\mathcal{M} \times T_p\mathcal{M} \to \mathbb{R}$ on the tangent space $T_p\gM$ at $p$.
For the tangent vectors $\xi, \zeta$ we write $\langle \xi, \zeta \rangle_p := g_p(\xi,\zeta)$ and $\|\xi\|_p := \sqrt{\langle \xi,\xi\rangle_p}$.
A 
\emph{geodesic} is a curve $\gamma:[0,1]\to\mathcal{M}$ that locally minimizes length.
The \emph{geodesic distance} between two points $p, q \in \gM$ is
\[
d_{\mathcal{M}}(p,q) \;:=\; \inf_{\gamma(0)=p,\,\gamma(1)=q}\; \int_0^1 \|\dot{\gamma}(t)\|_{\gamma(t)}\,dt .
\]

The \emph{exponential map} $\Exp_p : T_p\mathcal{M} \to \mathcal{M}$ maps a tangent vector $\xi$
to the endpoint of the geodesic starting at $p$ with initial velocity $\xi$:
$\Exp_p(\xi) := \gamma_{\xi}(1)$ where $\gamma_{\xi}(0)=p$, $\dot{\gamma}_{\xi}(0)=\xi$.
Its (local) inverse is the \emph{logarithm map} $\Log_p: \gM \to T_p\mathcal{M}$ satisfying
$\Exp_p(\Log_p(q)) = q$. These mappings are later used to obtain the Fr\'echet mean. 

\textbf{Quotients induced by symmetries.}
Neural network parameterizations often admit symmetries: many distinct parameter settings encode the same function.
A principled way to enforce symmetry-invariance is to work on a \emph{quotient manifold}.
Let $\overline{\mathcal{M}}$ be a (total) manifold with a smooth right action of a Lie group $\mathcal{G}$ so that
$
\Psi:\overline{\mathcal{M}}\times\mathcal{G}\to\overline{\mathcal{M}}$ and $(\bar{p},G)\mapsto \bar{p}\cdot G.
$ 
This action induces an equivalence relation $\bar{p}\sim \bar{q}$ if $\bar{q}=\bar{p}\cdot G$ for some $G\in\mathcal{G}$.
The quotient space is $\mathcal{M}:=\overline{\mathcal{M}}/\mathcal{G}$; we write $[\bar{p}]\in\mathcal{M}$ for the orbit
and $\pi:\overline{\mathcal{M}}\to\mathcal{M}$ for the projection $\pi(\bar{p})=[\bar{p}]$. 

\textbf{Vertical and horizontal spaces.}
At a point $\bar{p}\in\overline{\mathcal{M}}$, the \emph{vertical space}
$V_{\bar{p}}\overline{\mathcal{M}} \subset T_{\bar{p}}\overline{\mathcal{M}}$
is the tangent space to the orbit through $\bar{p}$, i.e.\ the set of infinitesimal symmetry directions.
If $\overline{\mathcal{M}}$ has a Riemannian metric $\bar{g}$ under which the action is by isometries (metric is symmetry invariant),
then we define the \emph{horizontal space} as the orthogonal complement w.r.t the Riemannian metric $\bar{g}$
\[
H_{\bar{p}}\overline{\mathcal{M}} \;:=\; \big(V_{\bar{p}}\overline{\mathcal{M}}\big)^\perp \subset T_{\bar{p}}\overline{\mathcal{M}}.
\]
Tangent vectors in the quotient space are set-valued and therefore inconvenient for numerical manipulation.
Conveniently, horizontal vectors provide representatives of tangent vectors on the quotient space:
each $\xi\in T_{[\bar{p}]}\mathcal{M}$ has a unique \emph{horizontal lift} $\bar{\xi}\in H_{\bar{p}}\overline{\mathcal{M}}$
with $d\pi_{\bar{p}}(\bar{\xi})=\xi$~\citep{absil2008optimization,boumal2023introduction}.

\textbf{Quotient distance as orbit alignment.}
If $\bar{g}$ is $\mathcal{G}$-invariant, it induces a well-defined metric $g$ on the quotient.
The resulting geodesic distance on $\mathcal{M}$ can be written as an \emph{orbit-minimum}:
\begin{equation}
\label{eq:quotient-distance}
d_{\mathcal{M}}\big([\bar{p}],[\bar{q}]\big)
\;=\;
\min_{G\in\mathcal{G}}\, d_{\bar{\mathcal{M}}}\big(\bar{p}, \bar{q}\cdot G\big).
\end{equation}
We will refer to a minimizer (a symmetry transformation that leads to the smallest geodesic distance)
\begin{equation}
\label{eq:alignment-def}
G^\star(\bar{p};\bar{q}) \in \arg\min_{G\in\mathcal{G}} d_{\bar{\mathcal{M}}}\big(\bar{p},\bar{q}\cdot G\big)^2
\end{equation}
as an \emph{alignment} (or \emph{gauge choice}) of $\bar{q}$ to $\bar{p}$, and write the aligned representative as
$\bar{q}^\star := \bar{q}\cdot G^\star(\bar{p};\bar{q})$.
Even when $\bar{q}$ and $\bar{q}^\star$ are different points in $\overline{\mathcal{M}}$, they represent the same quotient point:
$[\bar{q}]=[\bar{q}^\star]$. 
Alignment implies horizontal geodesics:
whenever $\bar{p}$ and $\bar{q}$ are aligned, $\Log_{\bar{p}}({\bar{q}})$ is horizontal (e.g., \citet{huckemann2010}), that is, the points are connected by a horizontal geodesic that is always perpendicular to orbits, i.e.\,$\dot{\gamma}(t) \in H_{\gamma(t)}\gM$. This will be a representation of the geodesic on the quotient space (which is an abstract space) that we can work with numerically. 

\subsection{Low-rank updates and their symmetries.}
A common PEFT primitive is a rank-$r$ update to a weight matrix. We consider updates of the form
\[
\Delta \mW \in \mathbb{R}^{d_1\times d_2},\quad \mathrm{rank}(\Delta \mW)=r\,,
\]
that admit several factorizations. We consider the following~\citep{hu2022lora, mishra2013fixed}: 

\textbf{(A) Standard factorization ($\mathrm{GL}(r)$ symmetry).}
$\Delta \mW$ is usually trained and stored as a factorization $\Delta \mW = \mG \mH^\top$ with
$\mG\in\mathbb{R}^{d_1\times r}_\star$ and $\mH\in\mathbb{R}^{d_2\times r}_\star$ ($_\star$ indicating full column rank).
Then, for any $\mA\in \mathrm{GL}(r)$, $(\mG\mA^{-1})(\mH\mA^\top)^\top = \mG\mH^\top$ and we define the equivalence relationship
\[
(\mG,\mH)\sim (\mG \mA^{-1},\, \mH \mA^\top).
\]

\vspace{-0.1in}
\textbf{(B) Polar factorization ($\mathrm{O}(r)$ symmetry).}
A rank-$r$ matrix admits an equivalent polar factorization
\begin{gather}
    \Delta \mW = \mU \mB \mV^\top
\end{gather}
where, $\mU\in \mathrm{St}(d_1,r),\; \mV\in \mathrm{St}(d_2,r),\; \mB\in \mathbb{S}_{++}(r)$, and $\mathrm{St}(d,r)=\{\mU\in\mathbb{R}^{d\times r}:\mU^\top \mU=\mI_r\}$ is the Stiefel manifold and
$\mathbb{S}_{++}(r)$ the SPD manifold of rank $r$.
This representation has the (smaller) orthogonal symmetry group:
for any $\mO\in \mathrm{O}(r)$,
\[
(\mU,\mB,\mV)\sim (\mU\mO,\; \mO^\top \mB \mO,\; \mV \mO),
\]
since $(\mU\mO)(\mO^\top \mB \mO)(\mV\mO)^\top = \mU\mB\mV^\top$.
This $\mathrm{O}(r)$-quotient viewpoint is particularly convenient computationally because geometric primitives like $\Exp$, and $\Log$ either admit exact expressions or accurate numerical routines~\citep{edelman1998geometry, mataigne2024log}.

\textbf{The quotient geometry of the polar factorization.}
We follow the construction from~\citet{mishra2013fixed}. 
On $\mathbb{S}_{++}$ we will use the affine-invariant metric:
\begin{align*}
g_\mB(\vzeta_\mB, \veta_\mB) = \langle \vzeta_\mB,\veta_\mB\rangle_\mB \;:=\; \Tr\!\big(\mB^{-1} \vzeta_\mB \mB^{-1} \veta_\mB \big)
\end{align*}
where $ \veta,\vzeta\in T_\mB\mathbb{S}_{++} = \{\veta=\veta^\top\}$ and with closed-form Riemannian exponential and logarithm:
\begin{subequations}
\begin{align}
    &\Exp_\mB(\veta)=\mB^{1/2}\exp\!\big(\mB^{-1/2} \veta \mB^{-1/2}\big)\mB^{1/2}\,,\\
&\Log_\mB(C)=\mB^{1/2}\log\!\big(\mB^{-1/2} C \mB^{-1/2}\big)\mB^{1/2}\,,
\end{align}
\end{subequations}
where $\exp_m(\cdot)$ denotes the matrix exponential.
For the Stiefel factors, we deviate from~\citet{mishra2013fixed} and use the canonical Stiefel metric
\begin{align*}
\langle \vzeta_\mU,\veta_\mU\rangle_\mU = \Tr\left(\vzeta_\mU^{\top}(\mI-\frac{1}{2}\mU \mU^\top) \veta_\mU \right)
\end{align*}
which has a better-behaved numerical routine for calculating its $\Log$ (using the algorithm from~\citet{mataigne2024log}).
\citet{edelman1998geometry} derived the exponential map: Given a point $\mU\in \mathrm{St}(n,r)$ and a tangent vector $\veta_\mU\in T_\mU\mathrm{St}(n,r)$, first create the compact QR factorization
$(\mI - \mU\mU^\top)\veta \;=\; \mQ\mR$, with $\mQ\in \mathrm{St}(n,r),\;\; R\in\mathbb{R}^{r\times r}$. Then, $\mA \;:=\; \mU^\top \veta$ is skew-symmetric, i.e., $\mA^\top=-\mA$.
With these definitions, the exponential mapping is
\begin{equation}
\label{eq:stiefel_exp_canonical}
\Exp_{\mU}(\veta)\;=\;
\begin{pmatrix} \mU & \mQ \end{pmatrix}
\exp\!\left(
\begin{pmatrix}
\mA & -\mR^\top\\
\mR & 0
\end{pmatrix}
\right)
\begin{pmatrix}
\mI_p\\
0
\end{pmatrix}
\end{equation}
and becomes particularly efficient when $r < n/2$ which is always the case for the LoRA adapters we will study---typically $(n=4096) \times (r=16)$ matrices.

The projection onto the horizontal space is given by
\begin{align*}
\eta^{\mathrm{hor}}=(\eta_\mU-\mU\vOmega,\; \eta_\mB-(\mB\vOmega-\vOmega \mB),\; \eta_V-\mV\vOmega),
\end{align*}
with the skew-symmetric $\vOmega$ as numerical solution to the equation
\begin{align*}
&\mB^{-1}\vOmega \mB + \mB\vOmega \mB^{-1} - \vOmega
=\frac12(\mV^\top \eta_\mV + \mU^\top \eta_\mU) - 
(\mB^{-1}\eta_\mB-\eta_\mB\mB^{-1})\,.
\end{align*}

\vspace{-0.15in}
\section{GeoMerge}
\label{sec:geomerge}
We formulate model merging as 
a (possibly quotient) Fréchet mean on a Riemannian manifold and derive a practical, symmetry-invariant 
computation for it via orbit alignment and geodesic updates. 
We work through an analytically solvable toy model of this section in~\cref{sec:toy-problem}.
\vspace{-0.1in} 
\subsection{Fr\'echet averaging}
\vspace{-0.1in} The central object in our framework is the \emph{Fr\'echet mean}: given points
$x_1,\dots,x_T$ on a metric space (or Riemannian manifold) $(\mathcal M,d)$ and weights
$w_i\ge 0$ with $\sum_i w_i=1$, a Fr\'echet mean is any minimizer of the \emph{Fr\'echet functional}
\vspace{-0.1in}
\begin{equation}
\mu^\star \in \arg\min_{\mu\in\mathcal M} F_{\mu}= \arg\min_{\mu\in\mathcal M} \frac12{\textstyle\sum_{i=1}^T} w_i d(\mu,x_i)^2.
\label{eq:frechet_functional}
\end{equation}
This definition is coordinate-free: it depends only on a notion of distance that reflects what it means
for two objects (models, distributions, adapters) to be ``close''. In Euclidean space $\mathcal{M}=\mathbb{R}^D$ and $d(\mu,x)=\|\mu-x\|_2$, and \eqref{eq:frechet_functional} reduces to weighted averaging; on curved spaces it produces an
\emph{intrinsic} average that respects the underlying geometry.
\vspace{-0.1in} 
\subsection{Why Fr\'echet means help in model merging}
Casting merging as \eqref{eq:frechet_functional} yields three benefits.
First, it cleanly separates \emph{what} we want (a geometry-respecting average) from \emph{how} we
compute it (distance approximations and optimization algorithms).
Second, it makes invariances explicit: if a symmetry acts by isometries, the induced distance is
symmetry-invariant and the merge is invariant \emph{by construction}.
Third, it unifies many existing merges as special cases obtained by different choices of
$\mathcal M$ and $d$ (or approximations thereof), and provides a principled way to derive new merges
by swapping in more appropriate geometry. Furthermore, we are not restricted to $d^2$ averages: one could take the Riemannian median ($d^1$), or even the Huber mean~\citep{lee2025hubermeansriemannian}, which mixes elements of both.

\subsection{Quotient manifolds imply symmetry invariance}
Many model parameterizations are \emph{not identifiable}: multiple parameter settings represent the same
intrinsic object due to architectural symmetries (permutations, scalings, low-rank gauge freedoms).
Let a Lie group $\gG$ act on a total space $(\overline{\mathcal M},\bar g)$ by \emph{isometries}. The intrinsic
space is the quotient $\mathcal M:=\overline{\mathcal M}/\gG$ whose points are equivalence classes
$[\bar p]=\{\bar p\cdot G \mid G\in \gG\}$.
The quotient distance can be written as an orbit-minimum:
\begin{equation}
d_{\mathcal M}\big([\bar\mu],[\bar p]\big)
=
\min_{G\in \gG}\ d_{\bar{\mathcal M}}\!\big(\bar\mu,\bar p\cdot G\big).
\label{eq:quotient_distance_orbit_min}
\end{equation}
Because the action is isometric, $d_{\mathcal M}$ is well-defined and independent of representatives.
Consequently, the Fr\'echet objective on the quotient,
$\sum_i d_{\mathcal M}([\bar\mu],[\bar p_i])^2$,
is invariant to reparameterizations within each orbit: the merge depends only on intrinsic content, not on
a particular gauge choice. This is the mathematical sense in which quotient GeoMerge is
\emph{symmetry-invariant by construction}.

\subsection{Generalization via geometry: a menu of aligned distances}
A key advantage of the geometric formulation is that we can choose distances that are
(i) appropriate for the object being merged and (ii) consistent with desired invariances.

\textbf{Distribution-space geometries.}
When models are viewed as distributions (or predictive conditionals), the Fisher metric induces the
Fisher-Rao geodesic distance; Fisher merging can be interpreted as a tractable approximation to
Fr\'echet averaging under this geometry (via a Gaussian/Laplace approximation and a quadratic
localization). Related divergences can serve as substitutes or bounds: for instance, symmetrized Jensen
divergences provide upper bounds on Fisher-Rao distance, offering alternative (though not always
tractable) objectives.


\textbf{Adapter-space geometries.}
For low-rank adapters, the intrinsic object is naturally a quotient, and we can equip the total
space with a product metric that respects the $O(r)$ gauge symmetry in the polar factorization. We use the canonical Stiefel metric for the Stiefel factors, and the affine-invariant metric for the SPD factor (see~\cref{sec:prelims}). 
\emph{Once a metric is chosen}, its induced distance is automatically
``aligned'' with the constraints and symmetries encoded by the manifold/quotient structure.

\subsection{Computing Fr\'echet means: the algorithmic recipe}
Computing \eqref{eq:frechet_functional} generally requires three ingredients:
(i) a way to evaluate or approximate $d_{\mathcal M}$, and/or
(ii) access to the exponential and logarithm maps $\Exp_\mu:T_\mu\mathcal M\to\mathcal M$ and
$\Log_\mu:\mathcal M\to T_\mu\mathcal M$, and
(iii) an optimization scheme.
A standard approach is Riemannian gradient descent on the Fr\'echet functional. When $\Log_\mu(x_i)$
is well-defined (e.g.\ within a geodesically convex neighborhood), the Riemannian gradient takes the
simple form
\begin{equation}
\mathrm{grad}\,F(\mu) = -{\textstyle\sum_{i=1}^T} w_i\, \Log_\mu(\theta_i),
\label{eq:frechet_grad}
\end{equation}
leading to the update
\begin{equation}
\mu_{t+1}
=
\Exp_{\mu_t}\!\left(
\alpha_t {\textstyle\sum_{i=1}^T} w_i\, \Log_{\mu_t}(\theta_i)
\right),
\label{eq:frechet_rg_update}
\end{equation}
with step size $\alpha_t>0$. If an exact closed-form for $d_{\mathcal M}$ is available and differentiable,
one can also differentiate \eqref{eq:frechet_functional} directly; otherwise \eqref{eq:frechet_rg_update}
provides a practical route whenever $\Exp/\Log$ (exact or approximate) are available.

\subsection{Quotient GeoMerge: alignment \& horizontal descent}
\begin{figure}
  \centering
  \sbox0{\includegraphics{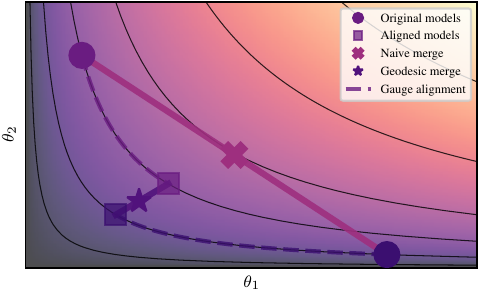}}%
  \begin{minipage}[c]{\wd0}\usebox0\end{minipage}\hfill
  \begin{minipage}[c]{\dimexpr\linewidth-\wd0-1em\relax}
    \caption{\emph{Geodesic merging in a toy two-parameter setting with a scaling symmetry.} Two checkpoints ($\bullet$) represent distinct models, each with infinitely many gauge-equivalent representatives along its symmetry orbit (dashed). Naïve ambient averaging yields a merge ($\cross$) that depends on the chosen representatives and can drift to a different orbit. GeoMerge instead aligns orbits ($\blacksquare$), then averages intrinsically along a horizontal geodesic, yielding a symmetry-consistent merge ($\star$).} \label{fig:geomerge}
  \end{minipage}
  \vspace{-2ex}
\end{figure}

\textbf{Intuition.} ~\cref{fig:geomerge} provides a schematic for quotient GeoMerge: the same intrinsic model/update can admit multiple equivalent representatives, and naïvely merging unaligned representatives can produce a result that is not representative of any sensible intrinsic average. Quotient GeoMerge resolves this by first aligning each input via the orbit-minimization step (selecting the best symmetry transformation), then performing the averaging using only the resulting horizontal directions. 

\textbf{Algorithm.} Practically, we implement quotient Fr\'echet descent by
alternating:
\begin{enumerate}
\item \textbf{Alignment (orbit minimization).} For a current representative $\bar\mu_t\in\bar{\mathcal M}$,
choose alignments
\begin{equation}
G_i^\star(\bar\mu_t)\in\arg\min_{G\in \gG}\ \tfrac12\, d_{\bar{\mathcal M}}\!\big(\bar\mu_t,\bar \theta_i\cdot G\big)^2,\quad \bar \theta_i^\star(\bar\mu_t):=\bar \theta_i\cdot G_i^\star(\bar\mu_t).
\label{eq:alignment_step}
\end{equation}
\item \textbf{Intrinsic averaging in the total space.} Compute total-space logarithms
$\eta_i:=\Log_{\bar\mu_t}(\bar \theta_i^\star(\bar\mu_t))\in T_{\bar\mu_t}\bar{\mathcal M}$, which is guaranteed to be horizontal (in orthogonal complement of the group orbit directions) at $\bar\theta_i^\star(\bar\mu_t)$, and update
\begin{equation}
\bar\mu_{t+1}=\Exp_{\bar\mu_t}\!\left(\alpha_t {\textstyle\sum_{i=1}^T} w_i\, \eta_i\right).
\label{eq:quotient_geomerge_update}
\end{equation}
\end{enumerate}
Intuitively, alignment removes gauge mismatch so that logarithms compare like-with-like, and as it is horizontal
we ignore directions that correspond purely to changing representatives rather than changing
the intrinsic quotient point. We provide concrete implementation details in~\Cref{sec:implementation}.

\section{Relation of GeoMerge to Fisher Merging}
\label{sec:relations}
\label{sec:fisher_geo}
We connect \emph{Fisher merging} \citep{matena2022fisher} 
 to GeoMerge by showing how it arises as tractable approximations to the Fréchet objective with an information-geometric distance.
Given inputs $\{x_j\}_{j=1}^N$ and a conditional model $p_\theta(y\mid x)$, the estimated diagonal Fisher at parameters $\theta$ is
\begin{equation*}
\label{eq:diag-fisher}
\widehat{F}_\theta
\;:=\;
\frac{1}{N}{\textstyle\sum_{j=1}^N}
\mathbb{E}_{y\sim p_\theta(y\mid x_j)}
\Big[
\big(\nabla_\theta \log p_\theta(y\mid x_j)\big)^{\odot 2}
\Big]
\;,
\end{equation*}
where $(\cdot)^{\odot 2}$ denotes elementwise squaring. 
Given models $\theta_1,\dots,\theta_T$ with corresponding diagonal Fishers $\widehat{F}_{\theta_i} \in \sR^D$,
Fisher merging computes a per-coordinate precision-weighted average:
\begin{equation*}
\label{eq:fisher-merge}
\theta_{\mathrm{Fisher}}^{\star}
\;=\;
\left({\textstyle\sum_{i=1}^T} \big(\widehat{F}_{\theta_i} \odot \theta_i\big)\right)
\oslash \left({\textstyle\sum_{i=1}^T}\widehat{F}_{\theta_i} \right),
\end{equation*}
with $\odot$ ($\oslash$) denoting elementwise multiplication (division).
This weighting is motivated probabilistically, by a Gaussian approximation to each model's posterior \citep{matena2022fisher}. 
The geometric lens of GeoMerge provides us a complementary view: \emph{Fisher merging arises by replacing an analytically intractable Fr\'echet objective under the Fisher-Rao geometry with a tractable local quadratic surrogate.}

Concretely, view each checkpoint $\theta_i$ as specifying an (approximate) Gaussian distribution over parameters $q_i(\vartheta)\;\approx\;\mathcal{N}\!(\theta_i,\;\smash{F_i^{-1}})$,
where $F_i$ denotes (an estimate of) the Fisher at $\theta_i$.
Consider the statistical manifold $\mathcal{Q}$ of Gaussians equipped with the Fisher information metric (the canonical metric in information geometry \citep{amari2000infogeo}), whose induced geodesic distance we denote by $d_{\mathrm{FR}}$ (Fisher-Rao distance).
A more principled merge is then the Fr\'echet mean of $\{q_i\}_{i=1}^T$:
\begin{equation}
q^\star \in \arg\min_{q\in\mathcal{Q}}\;\frac12{\textstyle\sum_{i=1}^T} d_{\mathrm{FR}}(q,q_i)^2.
\label{eq:fr_gaussian_objective}
\end{equation}
While $d_{\mathrm{FR}}$ between general Gaussians is not analytically available, it admits simple closed forms on two important submanifolds:
(i) for fixed covariance, it reduces to a Mahalanobis distance in the mean; and
(ii) for fixed mean, it reduces to a standard SPD distance in the covariance/precision (equivalently, an affine-invariant/log-Euclidean form).

\textbf{An upper bound that yields Fisher merging.}
Restrict the candidate to the Laplace family $q(\vartheta)=\mathcal{N}(\theta,F^{-1})$ with free mean $\theta$ and (possibly) free precision $F$.
For each $i$, introduce the intermediate Gaussian $\tilde q_i(\vartheta) := \mathcal{N}(\theta,\;F_i^{-1})$,
which shares covariance with $q_i$ and shares mean with $q$.
By the triangle inequality for $d_{\mathrm{FR}}$ and the elementary bound $(a+b)^2\le 2(a^2+b^2)$, we obtain (see, e.g. ~\citet{pinele2020fisherrao} for derivations of these distances)
\begin{equation*}
d_{\mathrm{FR}}\!\big(q_i,q\big)^2
\le 2\,d_{\mathrm{FR}}\!\big(q_i,\tilde q_i\big)^2 + 2\,d_{\mathrm{FR}}\!\big(\tilde q_i,q\big)^2
= 2\,(\theta-\theta_i)^\top F_i(\theta-\theta_i) + 2\,d_{\mathrm{SPD}}\!\big(F_i,F\big)^2,
\label{eq:fisherrao_upper_bound}
\end{equation*}
where $d_{\mathrm{SPD}}$ denotes the canonical geodesic distance on the $\mathrm{SPD}$ manifold, induced by the affine-invariant metric.
Crucially, the first term in the equation above is a \emph{quadratic} (Mahalanobis) distance in parameter space, and the second term depends only on the choice of $F$ (not on $\theta$).

If we are just concerned with a pointwise estimate and simply drop the covariance SPD term, then minimizing \eqref{eq:fr_gaussian_objective} can be approximated by the quadratic surrogate
\begin{equation}
\theta^\star \;\in\;\arg\min_{\theta}\;{\textstyle\sum_{i=1}^T} (\theta-\theta_i)^\top F_i(\theta-\theta_i).
\label{eq:quad_surrogate}
\end{equation}
The objective \eqref{eq:quad_surrogate} is strictly convex when $\sum_i F_i \succ 0$ and has the closed-form minimizer
\begin{equation}
\theta^\star
=\left({\textstyle\sum_{i=1}^T} F_i\right)^{-1}\left({\textstyle\sum_{i=1}^T} F_i\,\theta_i\right),
\label{eq:fisher_closed_form}
\end{equation}
which is exactly Fisher merging (with $F_i$ replaced by the chosen Fisher approximation, e.g.\,diagonal).

\section{GeoMerge: High-Rank} \label{sec:high_rank}

Some LoRA merging methods first embed rank-$r$ adapters into a larger rank budget and then apply
a merge rule in the higher-dimensional representation. E.g., KnOTS aligns task updates in
an SVD-derived coordinate system, while Core Space builds shared left/right bases before merging
the induced cores~\citep{stoica2025knots, panariello2025corespace}. From our perspective, these methods
combine two choices: a rank-increasing lift and an averaging rule. GeoMerge works as an averaging rule: a Fréchet mean on a quotient manifold. We thus need a quotient-compatible
lift from $\mathcal M_r$ to $\mathcal M_R$, where $R>r$ is the target rank budget.
Let
\[
\theta_t=[U_t,B_t,V_t]\in \mathcal M_r,\qquad
\mathcal M_r =
\bigl(\mathrm{St}(d_{\mathrm{out}},r)\times \mathrm{SPD}(r)\times
\mathrm{St}(d_{\mathrm{in}},r)\bigr)/O(r),
\]
and write the dense rank-$r$ update as $\Delta_t=U_tB_tV_t^\top$. A high-rank GeoMerge lift is a map
\[
\mathcal L_t:\{\theta_s\}_{s=1}^T \longmapsto
L_t=[\widehat U_t,\widehat B_t,\widehat V_t]\in \mathcal M_R,
\]
followed by the same quotient Fréchet objective as before: $
\mu_R^\star \in \arg\min_{\mu\in \mathcal M_R}
\frac{1}{2}\sum_{t=1}^T w_t\,d_{\mathcal M_R}(\mu,L_t)^2.
$
We argue that the lift should satisfy two main conditions. First, it must be defined on quotient points rather than on
arbitrary LoRA coordinates. Second, it must use the other task adapters, since
we want the added rank to utilize cross-task structure. 

The range of possible lifts is rather broad and we make no claims as to optimality; we present the simplest lift we could think of that satisfies the two criteria above. We leave a more thorough investigation of better choices for the lift to future work.

We choose orthonormal
complements $U_t^\perp\in\mathbb R^{d_{\mathrm{out}}\times (R-r)}$ and
$V_t^\perp\in\mathbb R^{d_{\mathrm{in}}\times (R-r)}$, and set $
\widehat U_t=[U_t,U_t^\perp],\,
\widehat V_t=[V_t,V_t^\perp].$
We define the projectors
$
P_t^U=I-U_tU_t^\top$ and $P_t^V=I-V_tV_t^\top.
$
The task-conditioned residual is
$
R_t=\sum_{s\neq t} P_t^U \Delta_s P_t^V .
$
We then take the leading paired singular directions
$
R_t=\widetilde U_t\Sigma_t\widetilde V_t^\top
$
and use them as the columns of $U_t^\perp$ and $V_t^\perp$.  We provide details on how we avoid instantiating the full dense matrix in~\cref{sec:non_dense}. The added rank is quotient-compatible:
$\Delta_s$, $P_t^U$, and $P_t^V$ are invariant to the $O(r)$ gauge of the input factors, while any sign or
rotation ambiguity in singular spaces is absorbed by the $O(R)$ quotient.

The lifted SPD factor can be picked in many different ways, but we keep it as simple as possible
\[
\widehat B_t=
\begin{bmatrix}
B_t & 0\\
0 & cI_{R-r}
\end{bmatrix},
\qquad
c=\left(\prod_{s=1}^T \lambda_{\min}(B_s)\right)^{1/T}.
\]
Scalar $c$ is a conservative SPD filler: it keeps $L_t$ in the fixed-rank manifold
$\mathcal M_R$ without disturbing the lift in a large way.

\section{Experiments}
\label{sec:experiments}
\begin{table}[t]
\centering
\caption{Merged-model performance on vision tasks for Vit-B/32 in normalized accuracies (\%).}
\resizebox{\textwidth}{!}{
\begin{tabular}{llccccccccc}
\toprule
\textbf{Method} & \textbf{Space} & \textbf{Cars} & \textbf{DTD} & \textbf{EuroSAT} & \textbf{GTSRB} & \textbf{MNIST} & \textbf{RESISC} & \textbf{SUN397} & \textbf{SVHN} & \textbf{Avg} \\
\midrule
TA & Full & 81.97 & 73.72 & 48.97 & 42.24 & 53.12 & 71.50 & 97.46 & 41.25 & 63.78  \\
\midrule
\multirow{3}{*}{TIES} & Full & 82.37 & 72.72 & 49.91 & 36.62 & 57.16 & 69.38 & 96.92 & 44.56 & 63.70  \\
& KnOTS  & 83.75 & 74.45 & 50.36 & 47.31 & 67.01 & 71.79 & 96.51 & 50.64 & 67.73  \\

 & Core & 84.74 & 76.46 & 52.19 & 50.41 & 67.36 & 71.21 & 96.45 & 50.18 & 68.63  \\
\midrule
\multirow{3}{*}{Iso-C} & Full        & 80.16 & 83.03 & 51.44 & 74.76 & 70.72 & 79.89 & 98.66 & 50.20 & 73.60 \\
& KnOTS   & 80.33 & 79.29 & 57.50 & 67.60 & 65.63 & 79.54 & 99.26 & 46.62 & 71.97  \\

 & Core    & 83.35 & 84.30 & 50.13 & 81.97 & 71.07 & 83.46 & 99.17 & 53.90 & 75.92 \\
 \midrule
\multirow{3}{*}{TSV+Iso-C} & Full  & 79.38 & 80.38 & 57.99 & 65.64 & 64.22 & 79.74 & 98.59 & 46.49 & 71.55 \\
& KnOTS & 80.81 & 83.03 & 58.25 & 74.34 & 67.66 & 79.69 & 98.54 & 49.86 & 74.02  \\

 & Core & 82.98 & 85.12 & 50.95 & 84.25 & 71.14 & 84.39 & 99.06 & 53.53 & 76.43\\
\midrule
 \multirow{2}{*}{\textbf{Ours}}
& TSV+Iso-C on our lift   & 82.47 & 84.21 & 53.75 & 81.96 & 72.48 & 79.88 &  99.26 &  53.72 & 75.97  \\

 & \textbf{GeoMerge+lift}    & 83.48 & 84.30 & 53.27 & 82.14 & 75.45 & 82.22 & 98.76 & 57.18 & \textbf{77.10} \\
\bottomrule
\end{tabular}}\label{tab:per-task-vitb}
\end{table}

While we see our main contribution as conceptual, we empirically validate our approach 
for proof of concept.
%
\noindent \textbf{Benchmarks.}
Following experimental setup from prior work~\citep{stoica2025knots}, we consider eight LoRA-adapted specialists derived from the same base ViT-B/32 model~\citep{dosovitskiy-etal-2020}, where each was fine-tuned on a different vision dataset: Cars~\citep{krause20133d}, DTD~\cite{cimpoi2014describing}, EuroSAT~\citep{helber2019eurosat}, GTSRB~\citep{stallkamp2011german}, MNIST~\citep{lecun1998mnist}, RESISC~\citep{cheng2017remote}, SUN397~\citep{xiao2016sun}, and SVHN~\citep{netzer2011reading}.
(We provide language tasks results in~\cref{sec:nli}).
%
%
We use Task Arithmetic after merging~\citep{ilharco2023taskarith}. 
\noindent \textbf{Baselines.} Knots and Core are traditionally used in conjunction with merging methods such as TIES, TSV and Iso-C~\citep{yadav2023ties, gargiulo2025tsv, marczak2025isoc} 
so we use those approaches as baselines.
\noindent \textbf{Metrics.}
We report \emph{normalized accuracy} on each dataset, defined per task as merged model accuracy divided by corresponding specialist accuracy.
We summarize results using average normalized accuracy across tasks.
\noindent \textbf{Results.}
\Cref{tab:per-task-vitb} summarizes the results on ViT-B/32. We outperform both Knots and Core, reaching new state-of-the-art performance on this benchmark, without the benefit of the vast literature around Euclidean weight averaging. As an ablation we also include the best performing Euclidean averaging method on our lift (after aligning the lifted adapters). The ablation confirms that the performance gains do not solely come from our lift, since we actually underperform the Core lift.

\section{Remarks, Limitations \& Future Work}
While KnOTS operates on the full weight matrix space, i.e., directly on the $\Delta \mW$s, and hence could reasonably be expected to be as fully symmetry invariant as it is possible to be, we posit that the difference in performance comes from the fact that the underlying geometry, once the symmetries are fully accounted for, is non-Euclidean and as such has curvature corrections not captured by KnOTS.
While performance of GeoMerge is good, it is unclear what the best lifting procedure should be. This could potentially limit the widespread applicability of our approach and merits further study.
Most merging methods in a PEFT setting use one of TIES or DARE-TIES as an add-on to a baseline merging/alignment procedure. This is not yet integrated into our framework. TIES and DARE-TIES, since they rely on parameter magnitudes and signs, are inherently coordinate-dependent procedures, whereas we take a completely coordinate-agnostic perspective. We leave this integration to future work.
While we believe the main contribution of this paper is conceptual, computational efficiency could still be improved, and might limit widespread usability.


\section{Conclusion}
GeoMerge proposes an alternative, geometric view on model merging by treating it as computing a Fréchet mean under an explicit geometry, rather than an arbitrary average in parameter space. This perspective directly addresses architectural and parameterization symmetries: when models live on orbits of equivalent representations. By operating on the appropriate manifold via orbit alignment and geodesic updates, our merges are symmetry-consistent by construction and grant access to the deeper geometric structure of parameter space. Potentially opening up a wide range

Our framework also yields practical algorithms and connections to prior work. We show how Fisher-weighted merging arises as a tractable approximation to an information-geometric Fréchet objective, and we instantiate geometric merging for LoRA adapters where symmetries are unavoidable. Empirical  evidence supports the effectiveness of our proposed method.


\medskip

{
\small
  \bibliographystyle{abbrvnat}
  \bibliography{references.bib}
}

\clearpage

\appendix
\section*{Appendix}
\section{GeoMerge: A Toy Problem}
We analyze a minimal two-layer linear model with a scaling symmetry to make the symmetry-induced failure modes of naïve parameter averaging concrete and to illustrate GeoMerge as quotient/geodesic merging in closed form.

\paragraph{A two-layer linear network with a scaling gauge.}
We start with the smallest setting where a continuous architectural symmetry already makes
``parameter-space averaging'' ill-posed.
Consider a scalar two-layer linear network
\begin{equation}
    f_{\theta}(x) \;=\; \theta_2 \theta_1 x,
    \quad \theta=(\theta_1,\theta_2)\in \overline{\mathcal M}:=(\mathbb R^\star)^2
\end{equation}
where $\mathbb R^\star:=\mathbb R\setminus\{0\}$.
This parametrization has a multiplicative gauge symmetry:
for any $a\in \mathbb R^\star$,
\begin{equation*}
    (\theta_1,\theta_2)\cdot a \;:=\; (a\,\theta_1,\; \theta_2/a),
\end{equation*}
and $f_{\theta\cdot a}\equiv f_\theta$ since the \emph{predictor}
$\beta(\theta):=\theta_1\theta_2$ is invariant under the action.
Thus the intrinsic object is the orbit $[\theta]\in \mathcal M:=\overline{\mathcal M}/\mathbb R^\star$.

\paragraph{A symmetry-invariant metric (scalar specialization of the $\mG\mH^\top$ metric).}
To respect the scaling symmetry, we equip $\overline{\mathcal M}$ with the
scalar version of the symmetry-invariant metric proposed for $(\mG,\mH)$-type parameters by~\citet{dasilva2025hide}:
\begin{align}
    g_{(\theta_1,\theta_2)}\!\big[(\eta_1,\eta_2),(\zeta_1,\zeta_2)\big]
    \;:=\;
    \theta_2^{2}\,\eta_1\zeta_1 \;+\; \theta_1^{2}\,\eta_2\zeta_2,
    \label{eq:toy_metric}
\end{align}
with $(\eta_1,\eta_2),(\zeta_1,\zeta_2)\in T_\theta\bar{\mathcal M}\cong\mathbb R^2$.
A direct calculation shows the group action acts by isometries under $g$ (so the quotient
$\mathcal M$ inherits a well-defined metric).

\paragraph{Vertical vs.\ horizontal directions.}
Let $a=\exp(s)$ and differentiate the group action at $s=0$ to obtain the vertical (orbit) direction $v(\theta)\;:=\;\left.\frac{d}{ds}\right|_{s=0}(\theta\cdot e^{s})
    =(\theta_1,\,-\theta_2)$.
The horizontal space is the $g$-orthogonal complement of $\mathrm{span}\{v(\theta)\}$, i.e.
$\eta=(\eta_1,\eta_2)$ is horizontal iff
\begin{equation}
    g_\theta\!\big[\eta, v(\theta)\big]
    \;=\; 0 \Longleftrightarrow\quad
    \frac{\eta_1}{\theta_1} \;=\; \frac{\eta_2}{\theta_2}.
    \label{eq:toy_horizontal}
\end{equation}
Intuitively, horizontal motion changes the intrinsic predictor $w=\theta_1\theta_2$,
while vertical motion changes only the representative (the gauge).

\paragraph{Horizontal geodesics make the predictor evolve linearly.}
In this toy geometry, geodesics admit a closed form. Writing $\gamma(t)=(\theta_1(t),\theta_2(t))$
with initial velocity $\dot\gamma(0)=(\eta_1(0),\eta_2(0))$, one obtains
\begin{align}
\begin{split}
    \theta_1(t) \;=\; \theta_1(0)\sqrt{1 + 2\,\eta_1(0)\,\theta_1(0)^{-1}t},
    \\
    \theta_2(t) \;=\; \theta_2(0)\sqrt{1 + 2\,\eta_2(0)\,\theta_2(0)^{-1}t}.
    \label{eq:toy_geodesic}
\end{split}
\end{align}
If the initial velocity is horizontal, \eqref{eq:toy_horizontal} implies
$\nicefrac{\eta_1(0)}{\theta_1(0)}=\nicefrac{\eta_2(0)}{\theta_2(0)}=:B$, and then the ratio $\nicefrac{\theta_1(t)}{\theta_2(t)}=\nicefrac{\theta_1(0)}{\theta_2(0)}$ is fixed, and
\begin{align}
\begin{split}
\beta(t):=\theta_1(t)\theta_2(t)
    \;=\; \theta_1(0)\theta_2(0)\big(1+2Bt\big).
    \label{eq:toy_predictor_linear}
\end{split}
\end{align}
Thus, once two points are \emph{gauge-aligned} so that a horizontal geodesic connects them,
the intrinsic interpolation is simply linear in predictor space.

\paragraph{Quotient distance reduces to predictor distance (after alignment).}
Because the action is isometric, the quotient distance can be written as an orbit-minimum:
\begin{equation*}
    d_{\mathcal M}\big([\theta],[\kappa]\big)
    \;=\; \min_{a\in\mathbb R^\star} d_{\overline{\mathcal M}}\big(\theta,\;\kappa\cdot a\big).
\end{equation*}
In this toy model, choosing $a$ so that $\theta$ and $\kappa\cdot a$ lie on the same horizontal geodesic
(i.e.\ they share the same ratio $\theta_1/\theta_2$ and lie in the same connected component/quadrant)
yields a closed-form distance that depends only on the invariant predictors:
\begin{equation}
    d_{\mathcal M}\big([\theta],[\kappa]\big)\;\propto\; \big|\,w(\theta)-w(\kappa)\,\big|
    \;=\;\big|\,\theta_1\theta_2-\kappa_1\kappa_2\,\big|.
    \label{eq:toy_quotient_distance}
\end{equation}

\paragraph{GeoMerge becomes ``average the predictors''.}
Given checkpoints $\theta^{(1)},\dots,\theta^{(N)}$, define $w_i:=w(\theta^{(i)})=\theta^{(i)}_1\theta^{(i)}_2$.
With \eqref{eq:toy_quotient_distance}, the quotient Fr\'echet objective becomes
\begin{align*}
    \mu^\star &= \arg\min_{[\mu]\in\mathcal M}\;\frac{1}{2}{\textstyle\sum_{i=1}^T} d_{\mathcal M}\big([\mu],[\theta^{(i)}]\big)^2
    \\ &=
    \arg\min_{w\in\mathbb R}\;\frac{1}{2}{\textstyle\sum_{i=1}^T} (w-w_i)^2,
\end{align*}
so the merged intrinsic predictor is simply $w^\star = \frac{1}{T}\sum_{i=1}^T w_i$.
Mapping back to parameters corresponds to choosing any representative $\mu=(\mu_1,\mu_2)$ with $\mu_1\mu_2=w^\star$;
this is exactly ``pick a gauge'' (e.g.\ enforce $\mu_1=\mu_2=\sqrt{|w^\star|}$ with a consistent sign choice).

\paragraph{Why this toy problem matters: a symmetry-induced pathology of Fisher/Euclidean averaging.}
Take two checkpoints that are \emph{the same function} but different representatives, e.g.\
$\theta^{(2)}=\theta^{(1)}\cdot (-1)=(-\theta^{(1)}_1,-\theta^{(1)}_2)$.
Na\"ive Euclidean averaging gives $(\theta^{(1)}+\theta^{(2)})/2=(0,0)$, which does not even lie in $\bar{\mathcal M}$
and corresponds to the zero predictor.
Moreover, in this toy setting, Fisher-weighted averaging exhibits the same failure mode:
because the Fisher can coincide for $\theta^{(1)}$ and $\theta^{(2)}$ while the parameters cancel,
the merged parameters collapse to $(0,0)$.
In contrast, GeoMerge works on the quotient where $[\theta^{(1)}]=[\theta^{(2)}]$ and hence
$d_{\mathcal M}([\theta^{(1)}],[\theta^{(2)}])=0$. We provide a quick derivation of the Fisher information matrix in~\cref{ap:fisher}.



\label{sec:toy-problem}
\section{Fisher Information Matrix for the Toy Model} \label{ap:fisher}

We consider the regression model defined by the function $f(x; \boldsymbol{\theta}) = \theta_1 \theta_2 x$. The observed data consists of $n$ pairs $\{(x_i, y_i)\}_{i=1}^n$, and we assume additive Gaussian noise with variance $\sigma^2$. The model is given by:
\begin{equation}
    y_i = \theta_1 \theta_2 x_i + \epsilon_i, \quad \epsilon_i \sim \mathcal{N}(0, \sigma^2)
\end{equation}
The parameter vector is $\boldsymbol{\theta} = [\theta_1, \theta_2]^\top$. The log-likelihood function $\ell(\boldsymbol{\theta})$ for the observations is:
\begin{equation}
    \ell(\boldsymbol{\theta}) = -\frac{n}{2}\log(2\pi\sigma^2) - \frac{1}{2\sigma^2} \sum_{i=1}^n (y_i - \theta_1 \theta_2 x_i)^2
\end{equation}

The score function is the gradient of the log-likelihood with respect to the parameters, $\nabla_{\boldsymbol{\theta}} \ell(\boldsymbol{\theta})$. The partial derivatives are:
\begin{align}
    \frac{\partial \ell}{\partial \theta_1} &= \frac{1}{\sigma^2} \sum_{i=1}^n (y_i - \theta_1 \theta_2 x_i)(\theta_2 x_i) \\
    \frac{\partial \ell}{\partial \theta_2} &= \frac{1}{\sigma^2} \sum_{i=1}^n (y_i - \theta_1 \theta_2 x_i)(\theta_1 x_i)
\end{align}

The Hessian matrix $\mathbf{H}$ consists of the second-order partial derivatives:
\begin{align}
    \frac{\partial^2 \ell}{\partial \theta_1^2} &= -\frac{1}{\sigma^2} \sum_{i=1}^n \theta_2^2 x_i^2 \\
    \frac{\partial^2 \ell}{\partial \theta_2^2} &= -\frac{1}{\sigma^2} \sum_{i=1}^n \theta_1^2 x_i^2 \\
    \frac{\partial^2 \ell}{\partial \theta_1 \partial \theta_2} &= \frac{1}{\sigma^2} \sum_{i=1}^n (y_i x_i - 2\theta_1 \theta_2 x_i^2)
\end{align}

The Fisher Information Matrix (FIM), $\mathcal{I}(\boldsymbol{\theta})$, is defined as the negative expectation of the Hessian matrix:
\begin{equation}
    \mathcal{I}(\boldsymbol{\theta}) = -E\left[ \mathbf{H} \right]
\end{equation}
We compute the expectation of the mixed partial derivative term using the relation $E[y_i] = \theta_1 \theta_2 x_i$:
\begin{equation}
    E\left[ \frac{\partial^2 \ell}{\partial \theta_1 \partial \theta_2} \right] = \frac{1}{\sigma^2} \sum_{i=1}^n (\theta_1 \theta_2 x_i^2 - 2\theta_1 \theta_2 x_i^2) = -\frac{1}{\sigma^2} \sum_{i=1}^n \theta_1 \theta_2 x_i^2
\end{equation}
Let $S_{xx} = \sum_{i=1}^n x_i^2$. The Fisher Information Matrix is therefore:
\begin{equation}
    \mathcal{I}(\boldsymbol{\theta}) = \frac{S_{xx}}{\sigma^2}
    \begin{bmatrix}
        \theta_2^2 & \theta_1 \theta_2 \\
        \theta_1 \theta_2 & \theta_1^2
    \end{bmatrix}
\end{equation}
Note that $\det(\mathcal{I}(\boldsymbol{\theta})) = 0$, indicating that the matrix is singular and the parameters $\theta_1$ and $\theta_2$ are unidentifiable as already pointed out. \label{sec:appendix}
\section{GeoMerge: Implementation Details}

This appendix gives the implementation-level details omitted from the main text. The central routine is
a quotient Fréchet mean: the same solver is used once at rank $r$ to obtain an anchor for gauge
alignment, and once at rank $R$ after the endpoint lift. We write a point in either manifold as
$p=[U,B,V]\in\mathcal M_k$, where $k=r$ or $k=R$.

The only subtlety is that the logarithm $\Log_p(q)$ must compare representatives in a common
gauge. Thus each Fréchet iteration first aligns every sample $q_i$ to the current iterate $p$, then
averages the resulting horizontal logarithms. For $q=[U_q,B_q,V_q]$, the $O(k)$ gauge action is
\[
q\cdot O=[U_qO,\;O^\top B_qO,\;V_qO],
\qquad O\in O(k).
\]
The alignment step approximately solves
\[
O^\star(p;q)
\in
\arg\min_{O\in O(k)}
d_{\bar{\mathcal M}_k}\bigl(p,q\cdot O\bigr)^2,
\]
where $\bar{\mathcal M}_k=\mathrm{St}(d_{\rm out},k)\times\mathrm{SPD}(k)\times
\mathrm{St}(d_{\rm in},k)$ is the total space. In practice, we use a Procrustes initialization followed
by a few horizontalization steps. The Procrustes initialization aligns the Stiefel factors:
$
O_0=\operatorname{polar}\!\left(U_q^\top U+V_q^\top V\right).
$
Given a current gauge $O$, we form the raw total-space logarithm from $p$ to $q\cdot O$ and solve
the horizontal projection equation for the skew-symmetric drift $\Omega$. We then update
\[
O\leftarrow O\exp(-\tau\Omega),
\]
for a small fixed number of inner iterations (we use 5 for the first iteration, 2 thereafter). The final logarithm returned to the Fréchet solver is the horizontal projection of the aligned raw logarithm.

\begin{algorithm}[H]
	\caption{Quotient Fréchet Mean with Orbit Alignment}
	\label{alg:quotient_frechet_mean}
	\begin{algorithmic}[1]
		\Require Quotient points $\{q_i=[U_i,B_i,V_i]\}_{i=1}^T\subset\mathcal M_k$, weights
		$\{w_i\}_{i=1}^T$, initializer $p_0$, step size $\alpha$, maximum iterations $N$, alignment
		iterations $A$, tolerance $\varepsilon$
		\Ensure Fréchet mean estimate $\mu\in\mathcal M_k$
		\State $p\leftarrow p_0$
		\For{$n=1$ to $N$}
		\For{$i=1$ to $T$}
		\State $O_i\leftarrow \operatorname{polar}(U_i^\top U_p+V_i^\top V_p)$
		\Comment{Procrustes initialization}
		\For{$a=1$ to $A$}
		\State $\tilde q_i\leftarrow q_i\cdot O_i$
		\State $\zeta_i\leftarrow \Log^{\rm total}_{p}(\tilde q_i)$
		\State $\Omega_i\leftarrow \operatorname{GaugeDrift}(p,\zeta_i)$
		\Comment{skew drift from horizontal equation}
		\State $O_i\leftarrow O_i\exp(-\tau\Omega_i)$
		\EndFor
		\State $\tilde q_i\leftarrow q_i\cdot O_i$
		\State $\eta_i\leftarrow \Pi^H_p\Log^{\rm total}_{p}(\tilde q_i)$
		\Comment{horizontal quotient log}
		\EndFor
		\State $\bar\eta\leftarrow \sum_{i=1}^T w_i\eta_i$
		\If{$\|\bar\eta\|_p<\varepsilon$}
		\State \Return $p$
		\EndIf
		\State $p\leftarrow \Exp_p(\alpha\bar\eta)$
		\EndFor
		\State \Return $p$
	\end{algorithmic}
\end{algorithm}

Here $\Pi^H_p$ denotes horizontal projection. For a raw tangent
$\zeta=(\zeta_U,\zeta_B,\zeta_V)$ at $p=[U,B,V]$, this projection has the form
\[
\Pi^H_p\zeta
=
\bigl(
\zeta_U-U\Omega,\;
\zeta_B-(B\Omega-\Omega B),\;
\zeta_V-V\Omega
\bigr),
\]
where $\Omega^\top=-\Omega$ is obtained from the horizontal gauge equation
\[
B^{-1}\Omega B+B\Omega B^{-1}-\Omega
=
\frac12\left(V^\top\zeta_V+U^\top\zeta_U\right)
-
\left(B^{-1}\zeta_B-\zeta_BB^{-1}\right).
\]
Thus $\operatorname{GaugeDrift}(p,\zeta)$ in Algorithm~\ref{alg:quotient_frechet_mean} is the
skew matrix $\Omega$ solving this equation. In the final line, $\Exp_p$ is implemented as the product
update on the total-space factors: a Stiefel update for $U$ and $V$, and the affine-invariant SPD
exponential for $B$.

Typically we use $\alpha = 1.0$ in our experiments since this seems to be stable. 

We next describe the algorithm for the rank-lifted procedure. After constructing lifted points $L_t\in\mathcal M_R$, it computes the final rank-$R$ quotient Fréchet mean.

\begin{algorithm}[t]
	\caption{Lifted GeoMerge}
	\label{alg:endpoint_lifted_geomerge}
	\begin{algorithmic}[1]
		\Require Rank-$r$ quotient points $\{\theta_t=[U_t,B_t,V_t]\}_{t=1}^T\subset\mathcal M_r$,
		target rank $R>r$, weights $\{w_t\}_{t=1}^T$
		\Ensure Rank-$R$ merged point $\mu_R\in\mathcal M_R$
		
		\State $c\leftarrow \left(\prod_{t=1}^T\lambda_{\min}(B_t)\right)^{1/T}$
		
		\For{$t=1$ to $T$}
		\State $\mathcal P_t\leftarrow \{s:s\neq t,\ a_{ts}\neq 0\}$
		
		\For{$s\in\mathcal P_t$}
		\State $A_{ts}\leftarrow U_s-U_t(U_t^\top U_s)$
		\State $C_{ts}\leftarrow V_s-V_t(V_t^\top V_s)$
		\Comment{applies $P_t^U$ and $P_t^V$ without forming them}
		\EndFor
		
		\State $A_t\leftarrow [A_{ts}]_{s\in\mathcal P_t}$,
		\quad
		$C_t\leftarrow [C_{ts}]_{s\in\mathcal P_t}$
		
		\State $M_t\leftarrow \operatorname{blkdiag}(a_{ts}B_s)_{s\in\mathcal P_t}$
		
		\State Thin span factorizations
		$
		A_t=E_t^UG_t,
		\qquad
		C_t=E_t^VH_t.
		$
		
		\State $K_t\leftarrow G_tM_tH_t^\top$
		\Comment{small core for the residual}
		
		\State Compute $K_t=\bar U_t\Sigma_t\bar V_t^\top$
		
		\State $U_t^\perp\leftarrow E_t^U\bar U_t$,
		\quad
		$V_t^\perp\leftarrow E_t^V\bar V_t$
		
		\State Keep the leading $R-r$ columns of $U_t^\perp$ and $V_t^\perp$
		
		\State $\widehat U_t\leftarrow [U_t,U_t^\perp]$,
		\quad
		$\widehat V_t\leftarrow [V_t,V_t^\perp]$
		
		\State $\widehat B_t\leftarrow
		\begin{bmatrix}
			B_t & 0\\
			0 & cI_{R-r}
		\end{bmatrix}$
		
		\State $L_t\leftarrow [\widehat U_t,\widehat B_t,\widehat V_t]\in\mathcal M_R$
		\EndFor
		
		\State $\mu_R\leftarrow
		\operatorname{FrechetMean}_{\mathcal M_R}(\{L_t\}_{t=1}^T;\{w_t\}_{t=1}^T)$
		\Comment{Algorithm~\ref{alg:quotient_frechet_mean}}
		
		\State \Return $\mu_R$
	\end{algorithmic}
\end{algorithm}

\subsection{Cayley Stiefel Updates and Approximate Logs}
\label{app:cayley_stiefel_updates}

The quotient Fréchet solver requires repeated Stiefel logarithm-like operations during orbit
alignment and mean updates. Exact canonical Stiefel logarithms are expensive in this setting, since
they are evaluated for every task, layer, and alignment iteration. We therefore use Cayley
pseudo-lifts and Cayley retractions for the Stiefel factors, while keeping the SPD factor using the exact
affine-invariant log/exp.

We use the notation of \citet{kaneko2013empiricalstiefel}. For a square matrix $M$, define
\begin{equation}
	\operatorname{sk}(M)
	:=
	\frac{1}{2}(M^\top-M).
	\label{eq:kaneko_sk_def}
\end{equation}
Thus $\operatorname{sk}^{-1}(M)$ denotes
$\bigl(\operatorname{sk}(M)\bigr)^{-1}$ when this inverse exists. Given
$X,Q\in \mathrm{St}(n,k)$, the full-Cayley pseudo-lift of
\citet{kaneko2013empiricalstiefel} is the ambient skew-symmetric matrix
\begin{equation}
	\widehat P_X^{-1}(Q)
	=
	\frac{1}{2}
	(Q-X)
	\operatorname{sk}^{-1}(X^\top Q)
	(Q-X)^\top
	\in \mathfrak{so}(n),
	\label{eq:kaneko_full_cayley_lift}
\end{equation}
provided that $\operatorname{sk}(X^\top Q)$ is nonsingular. This is the Cayley
``log'' used for the Stiefel factor in our quotient implementation.

The corresponding Cayley pseudo-retraction is
\begin{equation}
	\widehat P_X(\Omega)
	=
	\operatorname{Cay}(\Omega)X,
	\qquad
	\operatorname{Cay}(\Omega)
	=
	(I_n+\Omega)(I_n-\Omega)^{-1},
	\label{eq:kaneko_cayley_retraction}
\end{equation}
for $\Omega\in\mathfrak{so}(n)$.

Inside the quotient solver, after aligning the target representative to the current iterate, the raw
total-space log is assembled as
\begin{equation}
	\widetilde{\operatorname{Log}}^{\rm total}_{[U,B,V]}([U',B',V'])
	=
	\left(
	\widehat P_U^{-1}(U'),
	\operatorname{Log}^{\rm SPD}_{B}(B'),
	\widehat P_V^{-1}(V')
	\right),
	\label{eq:cayley_total_log}
\end{equation}
where the first and third components are the full-Cayley pseudo-lifts in
\eqref{eq:kaneko_full_cayley_lift}, and the middle component is the affine-invariant SPD logarithm.

For an horizontal tangent $(\Delta_U,\Delta_B,\Delta_V)$ at $[U,B,V]$, the update uses
Cayley retractions on the Stiefel factors and the affine-invariant exponential on the SPD factor:
\begin{equation}
	U^+
	=
	\operatorname{Cay}(\alpha \Omega_U)U,
	\qquad
	B^+
	=
	\operatorname{Exp}^{\rm SPD}_{B}(\alpha\Delta_B),
	\qquad
	V^+
	=
	\operatorname{Cay}(\alpha \Omega_V)V.
	\label{eq:quotient_cayley_update}
\end{equation}
Empirically, these approximations substantially reduce the cost of the repeated alignment
loop, with only a modest degradation relative to the exact Stiefel-log reference path. \label{sec:implementation}
\section{NLI Experiments}

\begin{table*}[t]
\centering
\caption{Merged-model performance on NLI tasks for Llama-3 8B in normalized accuracies, the ratio of the accuracy of the merged model on each dataset to that of the original LoRA adapters.}
\label{tab:nli}
\small
\setlength{\tabcolsep}{4.5pt}
\renewcommand{\arraystretch}{1.08}
\begin{tabular}{llccccccc}
\toprule
\multirow{2}{*}{\textbf{Method}} &
\multirow{2}{*}{\textbf{Framework}} &
\multicolumn{6}{c}{\textbf{Normalized Accuracy (\%)}} &
\multirow{2}{*}{\textbf{Avg.}} \\
\cmidrule(lr){3-8}
& & \textbf{SNLI} & \textbf{MNLI} & \textbf{SICK} & \textbf{QNLI} & \textbf{RTE} & \textbf{SciTail} & \\
\midrule

\multirow{1}{*}{TA} &
\textit{Standard}  &
93.57 & 95.28 & 87.96 & 68.71 & 100.0 & 96.73 & 90.38 \\
& \textit{GeoMerge}  & 
91.416 & 92.159 & 90.274 & 83.749 & 100.00 & 93.665 & 91.88 \\
\bottomrule
\end{tabular}
\end{table*}

Following prior work~\citep{stoica2025knots}, for language tasks we consider six LoRA-adapted specialists derived from the same Llama 3 8B base model~\citep{grattafiori2024llama},
each fine-tuned on a different NLI dataset: SNLI~\citep{bowman-etal-2015-large}, MNLI~\citep{mnli2018}, SICK~\citep{marelli-etal-2014-sick}, QNLI~\citep{wang-etal-2018-glue}, RTE~\citep{wang-etal-2018-glue}, and SciTail~\citep{khot-etal-2018-scitail}. 

\Cref{tab:nli} summarizes NLI merging results on Llama 3 8B: per-task normalized accuracy and average normalized accuracy.
GeoMerge outperforms the baseline in this setting. We did not run our full stack in this experimental setting due to computational limitations of our available Titan V GPU. We expect the full stack to perform considerably better,  since~\citet{panariello2025corespace} shows that restricting rank to the original rank $r$ has a severe effect on merged model performance. \label{sec:nli}

\section{Infrastructure Details}
\noindent \textbf{Compute Time and Resources.} Our merging algorithm runs on GPU and takes less than 1  hour on a machine with a Titan V GPU, with Intel(R) Xeon(R) W-2133 CPU 12 cores @ 3.60GHz.

In terms of compute time for the NLI datasets, on our machine, TA took 9s, KnOTS around 70 mins, GeoMerge at rank 16 with exact Exp/Log around 2hrs, and GeoMerge with Cayley retractions/logs at rank 16 around 60 mins. Lifted GeoMerge with Cayley retractions/log on the vision datasets took around 20 mins per run

\noindent \textbf{Dataset Licenses.} Among the datasets we use, we were able to determine the following licenses and/or usage permissions:
\begin{itemize}
    \item Cars~\citep{krause20133d} uses Creative Commons License.
    \item GTSRB~\citep{stallkamp2011german} uses Creative Commons License.
    \item EuroSAT~\citep{helber2019eurosat} uses MIT license.
    \item MNIST~\citep{lecun1998mnist} uses Gnu General Public License and elsewhere is listed under  Creative Commons Attribution-Share Alike 3.0 license.
    \item SUN397~\cite{xiao2016sun} is listed as ``for research purposes only''.
    \item SVHN~\cite{netzer2011reading} is listed as being ``for non-commercial use only''.
    \item SNLI~\citep{bowman-etal-2015-large} uses CC BY-SA 4.0 
    \item MNLI~\citep{mnli2018} appears to incorporate an aggregate of multiple licenses.
    \item SICK~\citep{marelli-etal-2014-sick} uses a Creative Commons Attribution-NonCommercial-ShareAlike license.
    \item QNLI~\citep{wang-etal-2018-glue} is derived from SQuAD, which in turn uses a CC BY-SA 4.0 license.
    \item RTE~\citep{wang-etal-2018-glue} appears to incorporate an aggregate of multiple licenses.
    \item SciTail~\citep{khot-etal-2018-scitail} uses an Apache 2.0 license.
\end{itemize} \label{sec:infra}
\section{Efficient High-Rank Lift}
We explain how we avoid instantiating the dense form of the completion basis used in
Section~\ref{sec:high_rank}. For a task $t$
\begin{equation}
	R_t
	=
	\sum_{s\neq t} P_t^U \Delta_s P_t^V,
	\qquad
	P_t^U = I-U_tU_t^\top,
	\qquad
	P_t^V = I-V_tV_t^\top,
	\label{eq:dense_residual}
\end{equation}
where $\Delta_s=U_sB_sV_s^\top$ is the rank-$r$ polar representation of task $s$. The completion
basis is then obtained from the leading paired singular directions of $R_t$. We now show how to
obtain the same subspaces without forming any dense matrix.

For each task $s\neq t$, define
\begin{equation}
	A_{ts}
	:=
	P_t^U U_s
	=
	U_s-U_t(U_t^\top U_s),
	\qquad
	C_{ts}
	:=
	P_t^V V_s
	=
	V_s-V_t(V_t^\top V_s).
	\label{eq:projected_peer_factors}
\end{equation}
Then
\begin{equation}
	P_t^U \Delta_s P_t^V
	=
	(P_t^U U_s)B_s(P_t^V V_s)^\top
	=
	A_{ts}B_sC_{ts}^\top,
\end{equation}
so
\begin{equation}
	R_t
	=
	\sum_{s\neq t} A_{ts}B_sC_{ts}^\top.
	\label{eq:low_rank_residual_sum}
\end{equation}

Stack all $s_i \neq t, i \in \{1, \ldots, T-1\}$
\begin{equation}
	A_t
	=
	\bigl[A_{s_1}, \ldots, A_{s_{T-1}}\bigr],
	\qquad
	C_t
	=
	\bigl[C_{s_1Sur}, \ldots, C_{s_{T-1}}\bigr],
\end{equation}
and let
\begin{equation}
	M_t
	=
	\operatorname{blkdiag}
	\bigl(
	B_{s_1},\ldots,B_{s_K}
	\bigr).
\end{equation}
Then
\begin{equation}
	R_t = A_tM_tC_t^\top.
	\label{eq:stacked_residual}
\end{equation}
Compute thin span factorizations
\begin{equation}
	A_t = E_t^U G_t,
	\qquad
	C_t = E_t^V H_t,
	\qquad
	(E_t^U)^\top E_t^U=I,
	\qquad
	(E_t^V)^\top E_t^V=I,
\end{equation}
Substituting into
\eqref{eq:stacked_residual} gives
\begin{equation}
	R_t
	=
	E_t^U K_t (E_t^V)^\top,
	\qquad
	K_t := G_tM_tH_t^\top.
	\label{eq:small_core_residual}
\end{equation}
The matrix $K_t$ has size at most $(T-1)r\times (T-1)r$.

Now take the small SVD
\begin{equation}
	K_t=\bar U_t\Sigma_t\bar V_t^\top.
\end{equation}
Then
\begin{equation}
	R_t
	=
	(E_t^U\bar U_t)\Sigma_t(E_t^V\bar V_t)^\top.
\end{equation}
Therefore the dense residual's nonzero singular directions are
\begin{equation}
	U_t^\perp = E_t^U\bar U_t,
	\qquad
	V_t^\perp = E_t^V\bar V_t.
	\label{eq:dense_free_completion_basis}
\end{equation}

This construction is exactly equivalent to forming $R_t$ and taking its SVD: the nonzero singular
values of $R_t$ and $K_t$ coincide, and their singular vectors are related by the isometries
$E_t^U$ and $E_t^V$. If singular values are repeated, equality is understood as equality of singular
subspaces, up to signs and orthogonal rotations inside degenerate subspaces. This ambiguity is
absorbed by the $O(R)$ quotient gauge.

Finally, the construction is quotient-defined. Under a peer gauge transformation
\begin{equation}
	(U_s,B_s,V_s)
	\mapsto
	(U_sQ_s,Q_s^\top B_sQ_s,V_sQ_s),
	\qquad Q_s\in O(r),
\end{equation}
one has
\begin{equation}
	A_{ts}\mapsto A_{ts}Q_s,
	\qquad
	C_{ts}\mapsto C_{ts}Q_s,
\end{equation}
and hence
\begin{equation}
	(A_{ts}Q_s)(Q_s^\top B_sQ_s)(C_{ts}Q_s)^\top
	=
	A_{ts}B_sC_{ts}^\top.
\end{equation}
The anchor projectors $U_tU_t^\top$ and $V_tV_t^\top$ are themselves invariant under the anchor
gauge. Thus $R_t$ and its completion subspaces depend only on the quotient points, not on the
chosen representatives. \label{sec:non_dense}
\newpage
\section*{NeurIPS Paper Checklist}

\begin{enumerate}

\item {\bf Claims}
    \item[] Question: Do the main claims made in the abstract and introduction accurately reflect the paper's contributions and scope?
    \item[] Answer: \answerYes{} 
    \item[] Justification: The reason that the main claims made in the abstract and introduction accurately reflect the paper's contributions and scope is because that is exactly one of the main purposes of the abstract and introduction. So, in turn, that was exactly one of our main objectives when writing the abstract and introduction. 

\item {\bf Limitations}
    \item[] Question: Does the paper discuss the limitations of the work performed by the authors?
    \item[] Answer: \answerYes{} 
    \item[] Justification: We have included a separate section entitled ``Remarks, Limitations \& Future Work'' in which we discuss the limitations of this work.

\item {\bf Theory assumptions and proofs}
    \item[] Question: For each theoretical result, does the paper provide the full set of assumptions and a complete (and correct) proof?
    \item[] Answer: \answerYes{} 
    \item[] Justification: We see this paper as making a primary theoretical contribution, supported by strong empirical evidence. We introduce and develop theory for Riemannian-based model merging, and we begin this by providing our definitions and assumptions. While we have not formulated the paper in terms of theorems and proofs (this would have been unnecessary), we have taken considerable efforts to explain our reasoning carefully, and to provide appropriate references when making use of a result that has been previously established. Where no reference is provided is because we assume the statement to be covered in standard mathematical references on differential and riemannian geometry.

    \item {\bf Experimental result reproducibility}
    \item[] Question: Does the paper fully disclose all the information needed to reproduce the main experimental results of the paper to the extent that it affects the main claims and/or conclusions of the paper (regardless of whether the code and data are provided or not)?
    \item[] Answer: \answerYes 
    \item[] Justification: Our intention in writing this paper is that if someone were to follow our recipe as we have described it---which we believe is entirely doable based on our provided description---then they will get essentially the same results.

\item {\bf Open access to data and code}
    \item[] Question: Does the paper provide open access to the data and code, with sufficient instructions to faithfully reproduce the main experimental results, as described in supplemental material?
    \item[] Answer: \answerNo{} 
    \item[] Justification: We intend to make our code available upon acceptance. The datasets we use are publicly available.

\item {\bf Experimental setting/details}
    \item[] Question: Does the paper specify all the training and test details (e.g., data splits, hyperparameters, how they were chosen, type of optimizer) necessary to understand the results?
    \item[] Answer: \answerYes{} 
    \item[] Justification: We follow experimental setups previously reported in the literature and with publicly available results. Details about our optimization and hyperparameters are reported in~\Cref{sec:implementation}.

\item {\bf Experiment statistical significance}
    \item[] Question: Does the paper report error bars suitably and correctly defined or other appropriate information about the statistical significance of the experiments?
    \item[] Answer: \answerNo{} 
    \item[] Justification: Due to computational constraints we were not able to provide these.

\item {\bf Experiments compute resources}
    \item[] Question: For each experiment, does the paper provide sufficient information on the computer resources (type of compute workers, memory, time of execution) needed to reproduce the experiments?
    \item[] Answer: \answerYes{} 
    \item[] Justification: Provided in~\Cref{sec:infra}
    
\item {\bf Code of ethics}
    \item[] Question: Does the research conducted in the paper conform, in every respect, with the NeurIPS Code of Ethics \url{https://neurips.cc/public/EthicsGuidelines}?
    \item[] Answer: \answerYes{} 
    \item[] Justification: This work conforms in every respect with the NeurIPS Code of Ethics.

\item {\bf Broader impacts}
    \item[] Question: Does the paper discuss both potential positive societal impacts and negative societal impacts of the work performed?
    \item[] Answer: \answerNA{} 
    \item[] Justification: This paper presents work whose goal is to advance the field of machine learning, specifically through the use of Riemannian geometry to improve model merging. While the broad field of machine learning itself has many potential societal consequences, we do not feel that there are any societal impacts specific to this work that need to be highlighted.
    
\item {\bf Safeguards}
    \item[] Question: Does the paper describe safeguards that have been put in place for responsible release of data or models that have a high risk for misuse (e.g., pre-trained language models, image generators, or scraped datasets)?
    \item[] Answer: \answerNA{} 
    \item[] Justification: This paper poses no such risks.

\item {\bf Licenses for existing assets}
    \item[] Question: Are the creators or original owners of assets (e.g., code, data, models), used in the paper, properly credited and are the license and terms of use explicitly mentioned and properly respected?
    \item[] Answer: \answerYes 
    \item[] Justification: All of the datasets and models we use are publicly available and have been described in prior work; we have, to the best of our knowledge, properly credited that prior work. We have also provided a section in the appendix that lists the names of the licenses we were able to identify for each of the datasets we used.

\item {\bf New assets}
    \item[] Question: Are new assets introduced in the paper well documented and is the documentation provided alongside the assets?
    \item[] Answer: \answerNA{} 
    \item[] Justification: The paper does not release new assets.

\item {\bf Crowdsourcing and research with human subjects}
    \item[] Question: For crowdsourcing experiments and research with human subjects, does the paper include the full text of instructions given to participants and screenshots, if applicable, as well as details about compensation (if any)? 
    \item[] Answer: \answerNA{} 
    \item[] Justification: This work included neither crowdsourcing nor research with human subjects.

\item {\bf Institutional review board (IRB) approvals or equivalent for research with human subjects}
    \item[] Question: Does the paper describe potential risks incurred by study participants, whether such risks were disclosed to the subjects, and whether Institutional Review Board (IRB) approvals (or an equivalent approval/review based on the requirements of your country or institution) were obtained?
    \item[] Answer: \answerNA{} 
    \item[] Justification: This work included neither crowdsourcing nor research with human subjects.

\item {\bf Declaration of LLM usage}
    \item[] Question: Does the paper describe the usage of LLMs if it is an important, original, or non-standard component of the core methods in this research? Note that if the LLM is used only for writing, editing, or formatting purposes and does \emph{not} impact the core methodology, scientific rigor, or originality of the research, declaration is not required.
    \item[] Answer: \answerNA{} 
    \item[] Justification: The core methodology does not use LLMs.

\end{enumerate}
\end{document}